\documentclass[10pt,twocolumn,letterpaper]{article}

\usepackage{cvpr}              
\usepackage[dvipsnames, table]{xcolor} 

\usepackage[dvipsnames]{xcolor}

\definecolor{cvprblue}{rgb}{0.21,0.49,0.74}
\usepackage[pagebackref,breaklinks,colorlinks,citecolor=cvprblue]{hyperref}

\usepackage{multirow}
\usepackage{blindtext}
\usepackage{tcolorbox}
\usepackage{graphicx} 
\usepackage{pifont}
\def\paperID{8579} 
\def\confName{CVPR}
\def\confYear{2024}

\title{CLIP-BEVFormer: Enhancing Multi-View Image-Based BEV Detector with Ground Truth Flow}

\author{Chenbin Pan$^{1,2}$\thanks{Corresponding to: \tt{chenbin.pan@us.bosch.com} } \ \  Burhaneddin Yaman$^{2}$ \ Senem Velipasalar$^{1}$ \ Liu Ren$^{2}$\\
$^{1}$Syracuse University \ \ \ \\ $^{2}$Bosch Research North America \& Bosch Center for Artificial Intelligence (BCAI)\\
{\tt\small \{cpan14, svelipas\}@syr.edu}, \ \ \ {\tt\small \{burhaneddin.yaman,liu.ren\}@us.bosch.com}
}

\begin{document}
\maketitle
\begin{abstract}
Autonomous driving stands as a pivotal domain in computer vision, shaping the future of transportation. Within this paradigm, the backbone of the system plays a crucial role in interpreting the complex environment. However, a notable challenge has been the loss of clear supervision when it comes to Bird's Eye View elements. To address this limitation, we introduce CLIP-BEVFormer, a novel approach that leverages the power of contrastive learning techniques to enhance the multi-view image-derived BEV backbones with ground truth information flow. We conduct extensive experiments on the challenging nuScenes dataset and showcase significant and consistent improvements over the SOTA. Specifically, CLIP-BEVFormer achieves an impressive 8.5\% and 9.2\% enhancement in terms of NDS and mAP, respectively, over the previous best BEV model on the 3D object detection task. 
\end{abstract}    
\section{Introduction}
\label{sec:intro}
Autonomous Driving Systems (ADS) have witnessed rapid advancements, revolutionizing the landscape of transportation \cite{STP3,chitta2022transfuser,Uniad,VAD,chen2023end,sparse4d,sparse4dv2,sparse4dv3}. The current state-of-the-art vision based ADS approaches heavily rely on Bird's Eye View (BEV) representations extracted from multi-view images to perceive and understand the surrounding environment. The BEV detector, which acts as the backbone of ADS, transforms multi-view images into a top-down view representation known as the BEV feature map \cite{philion2020lift,pan2020cross,fiery,bevformer,yang2023bevformer}. The effectiveness of the BEV detector significantly influences the success of perception tasks in autonomous driving, underscoring its pivotal role in enhancing the system's situational awareness \cite{bevformer,Uniad,VAD}.

However, in current BEV models such as BEVFormer \cite{bevformer}, the BEV encoding process and its information embedding quality are ensured by decoding specific 3D object information through a transformer decoder \cite{bevformer}. This dependency on the decoder's capability lacks a compelling force to align the produced BEV with the ground truth BEV. Additionally, decoding is executed using a set of initialized queries \cite{bevformer,detr}. In particular, through cross-attention and self-attention mechanisms \cite{vaswani2017attention}, the final informed queries align with ground-truth instances through a matching algorithm \cite{detr}. The decoding process resembles a black box and lacks interpretability, leaving uncertainty about which query corresponds to predicting which ground truth instance before decoding. Hence, there is a lack of a ground-truth perspective on the decoding process, such as how each ground truth instance interacts with others and the BEV feature map in the real scenarios.

In order to address the aforementioned limitations, we introduce CLIP-BEVFormer framework, which is composed of Ground Truth BEV (GT-BEV) module and Ground Truth Query Interaction (GT-QI) module.

In the absence of supervision on the BEV in previous models, the representation of object classes and boundaries is not guaranteed to align with the expressive nature of the ground truth BEV. This limitation imposes constraints on the perception capabilities of the model, leading to a reduced discriminative capacity. Hence, we introduce the GT-BEV module, a crucial component of our CLIP-BEVFormer framework, to enhance the quality of BEV generation. GT-BEV employs a ground-truth information flow (GT-flow) guidance during the BEV encoding phase. In particular, we implement a contrastive learning technique, as presented in CLIP \cite{clip}, to align the produced BEV with the ground truth BEV, ensuring explicit arrangement of BEV elements based on their class label, location, and boundary. This explicit element arrangement, guided by ground truth information, serves to enhance the perceptual abilities of the model, allowing for improved detection and differentiation of various objects on the BEV map.

The intricate interactions among ground truth instances on the BEV map have remained unexplored by previous models. The recovery of ground truth information from empty queries through self-attention and cross-attention mechanisms is limited in its interpretability, functioning as a black box exploration process and offering supervision only at the endpoint. To address this limitation, we introduce the Ground Truth Query Interaction (GT-QI) module, an integral component of our CLIP-BEVFormer framework. The GT-QI module injects GT-flow into the decoder during training, enriching the query pool and providing valuable learning insights into the decoding process. The incorporation of GT-flow into the decoding process facilitates interactions and communication among ground truth instances and between ground truth instances and the BEV map. By incorporating ground truth queries, the expanded query pool not only enhances the model's robustness but also augments its ability to utilize queries for detecting various objects within the source map.

Our innovative CLIP-BEVFormer, aiming to enhance the image-based BEV transformer with GT-flow guidance, improves both the BEV encoder process and the perception decoder process. It is a novel training framework that can be applied to any transformer and image based BEV detectors. Moreover, our method doesn't introduce any additional parameters and computations during inference stage, thus it maintains the efficiency of the original model. 

Our main contributions can be summarized as follows:
\begin{itemize}
    \item \textbf{CLIP-BEVFormer Framework:} We propose CLIP-BEVFormer, a pioneering training framework that enhances the BEV detector by integrating ground truth flow guidance into both BEV encoding and perception decoding processes. 
    \item \textbf{Superior Performance:} Through extensive experiments on challenging nuScenes dataset \cite{nuscenes}, we show that CLIP-BEVFormer consistently outperforms counterpart BEV detector methods on various tasks.
    \item \textbf{Generalization and Robustness:} Our model exhibits superior generalization and robustness, particularly in long-tail cases and scenarios involving sensor failures, ensuring heightened safety in autonomous driving.
    \item \textbf{Flexibility and Efficiency:} CLIP-BEVFormer showcase the flexibility of not necessitating a language model for training, relying on a simple MLP layer for substantial performance gains. CLIP-BEVFormer does not incur any extra computation time during inference time as its novel components are only employed during training time.
\end{itemize}

\section{Related Works}
\label{sec:related_work}
\subsection{Bird's Eye View Feature Generation}
BEV feature generation has recently gained a lot of interest for various downstream tasks as its holistic representation of the scene has been a success for various downstream tasks. Early works \cite{chen2017multi,philion2020lift,pan2020cross} leverages  ConvNets and inverse perspective mapping (IPM) for mapping features from perspective view to BEV view. More recently, transformers based architectures have been extensively studied for BEV feature generation \cite{yang2021projecting,chen2022persformer,fiery,beverse,bevformer}. While some of these works focuses on only spatial feature transformation \cite{chen2022persformer,yang2021projecting},  more recent works also incorporates temporal information for BEV generation. In particular,  BEVFormer \cite{bevformer} proposes a spatiotemporal transformer which employs spatial cross-attention to aggregate spatial features from multi-view camera images and  temporal self-attention to fuse history BEV features. 

BEV training is conducted by attaching downstream task heads to the generated BEV representation. 3D perception tasks such as 3D object detection and segmentation are two of the main downstream tasks for applications of BEV representations. The evaluation performance on downstream tasks serves as an indicator for the quality of BEV formation technique. 

\subsection{Vision-Language Models}
Vision-language models (VLMs) have shown great promise for learning good representations for variety of downstream tasks \cite{clip,singh2022flava,zhang2022contrastive,yuan2021florence,align,flamingo,pan2024vlp}. The success of VLMs have been driven by training transformers on large scale image-text pairs data collected from web using contrastive learning ~\cite{clip,align}. Among  VLMs, CLIP \cite{clip} which have been trained on 400 million pairs of data have shown remarkable zero-shot generalization on various image recognition tasks. The contrastive learning mechanism which maps image and text pairs to a joint embedding space has been pivotal for CLIP success. While VLMs generally show good performance on downstream tasks without any fine-tuning, prompting and fine-tuning techniques have also been employed for adaptation of VLMs to a new downstream task ~\cite{coop,adapter_clip}. 

While VLMs have been extensively explored and used in various domains, its application to BEV generation process is yet to be explored. A recent work  has explored using CLIP for BEV retrieval \cite{anonymous2023bevclip}. However, unlike our proposed approach, this work has no impact on BEV generation.

\subsection{Contrastive Learning in Computer Vision}
Contrastive learning has gained prominence in computer vision for self-supervised representation learning. Techniques like SimCLR \cite{simclr} and CLIP \cite{clip} demonstrate the power of contrastive learning in capturing meaningful representations from diverse data modalities. These approaches have primarily been applied to image and text domains.

However, the application of contrastive learning in the perception domain, particularly in the context of Bird's Eye View (BEV) detection, has been limited. Previous models \cite{clip-1,clip-2,clip-3,clip-4,clip-5}, including CLIP, often employ dual pathways for multi-modality incorporation and feature transfer learning. This dual-path approach necessitates the application of both pathways during inference, potentially impacting efficiency. In contrast to previous methodologies, our approach harnesses contrastive learning as a guiding mechanism to facilitate model parameter learning.

\section{Methodology}
\label{sec:methodology}
\begin{figure*}[bt!]
  \centering\includegraphics[width=0.9\linewidth]{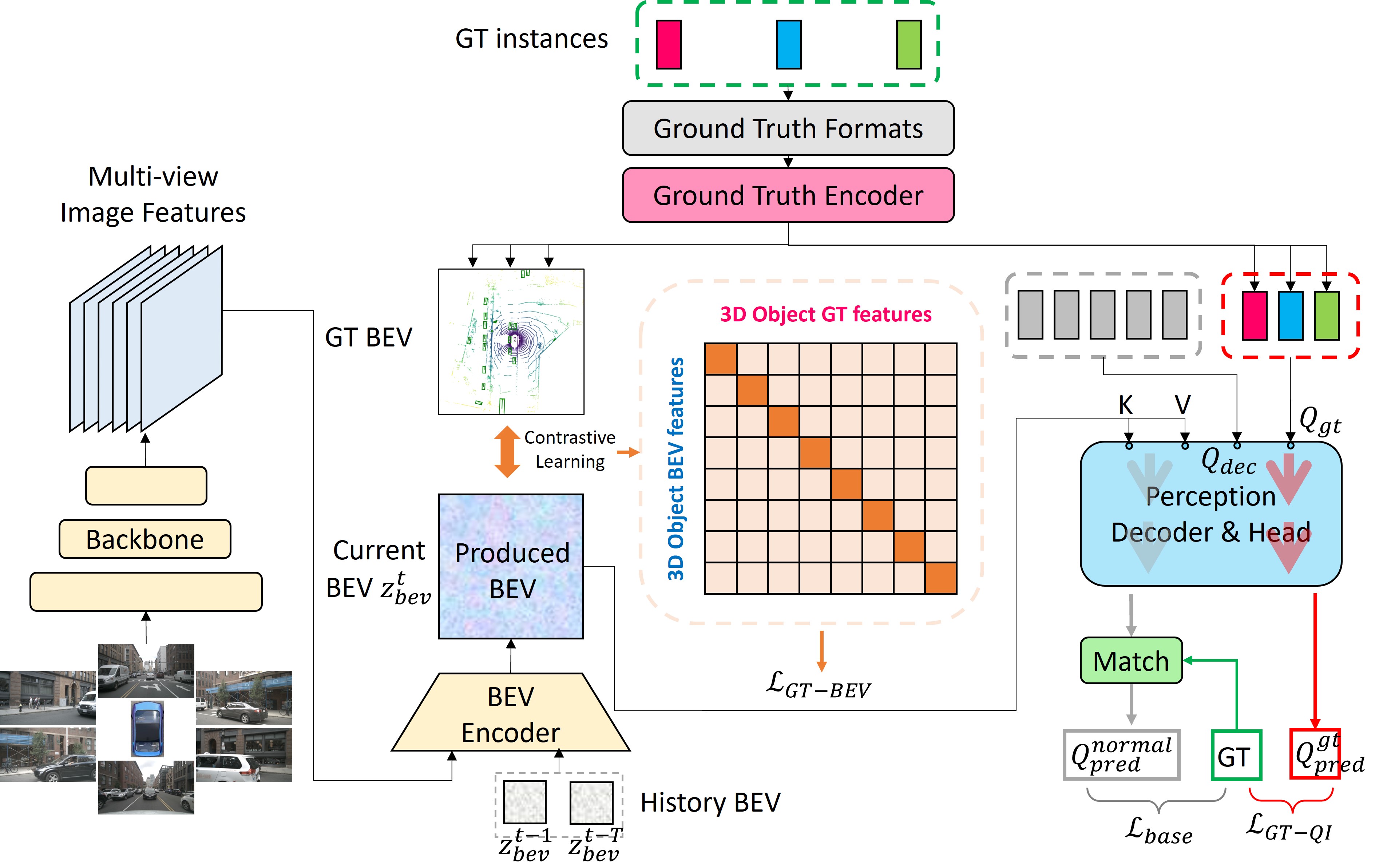}
   \caption{\textbf{The overview of CLIP-BEVFormer.} The architecture integrates two key modules: the Ground Truth BEV (GT-BEV) module employs a contrastive learning framework inspired by CLIP \cite{clip} to enrich the quality of BEV representations, while the Ground Truth Query Interaction (GT-QI) module introduces ground truth flow guidance into perception decoding processes. This integration leads to superior 3D object detection performance, as demonstrated in our extensive experiments on the challenging nuScenes dataset \cite{nuscenes}.}
   \label{fig:model}
   \vspace{-0.3cm}
\end{figure*}

We present the technical details of our proposed CLIP-BEVFormer illustrated in Fig.~\ref{fig:model}. 
The core innovation of CLIP-BEVFormer lies in its provision of a ground truth perspective for both the Bird's Eye View (BEV) encoding and perception query decoding processes, achieved through the integration of the Ground Truth BEV (GT-BEV) module and the Ground Truth Query Interaction (GT-QI) module, respectively.
We begin by examining the architecture of previous BEV detectors in Sec.~\ref{ssec:prev}. Subsequently, we delve into the specifics of the GT-BEV and GT-QI modules in Sec.~\ref{ssec:gt-bev} and Sec.~\ref{ssec:gt-query}, respectively. The training loss of our framework is outlined in Sec.~\ref{ssec:loss}.

\subsection{Preliminary}
\label{ssec:prev}
In the realm of image-based Bird's Eye View (BEV) detectors, the processing pipeline traditionally involves the utilization of multi-view camera images $\mathcal{X}_{views}$. These images undergo initial processing through an image backbone, followed by a BEV encoder to amalgamate pertinent image features into a unified top-down view BEV feature map $z_{bev} \in \mathbb{R}^{H_b \times W_b \times C}$. Based on transformer architecture, a set of decoder queries $q_{dec}$ is initialized to extract information from the BEV feature map~\cite{bevformer}. These queries are responsible for decoding perceptual information specific to objects within the receptive view based on the acquired information. The informed decoder queries are then directed to a perception head $Head$, generating detection results through a matching algorithm $Match$~\cite{detr}. Each query serves as a prediction for a matched instance or an empty instance, determined through the matching algorithm \cite{detr}. Following the matching process, the final perception loss is computed by applying a perceptual loss function between the query predictions and the ground truth instances. 
More formally, the process can be expressed as Eq.~(\ref{eq:prev-bev-det}):

\begin{equation}
\begin{aligned}
z_{bev} &= BEVEnc(\mathcal{X}_{views}),\\
q_{dec}^{'} &= Dec(z_{bev}, q_{dec}),\\
q_{pred} &= Match(Head(q_{dec}^{'}) , y),\\
\mathcal{L} &= \mathcal{L}_{perc}(q_{pred} , y),
\end{aligned}
\label{eq:prev-bev-det}
\end{equation}

where $BEVEnc$ and $Dec$ denote the BEV encoder and perception decoder respectively, $y$ denotes the ground truth of the perception task, and $q_{pred}$ denotes the query predictions after matching.

\subsection{Ground Truth BEV}
\label{ssec:gt-bev}

In the absence of supervision on the BEV in previous models, the accurate representation of object classes and boundaries is not guaranteed to align with the expressive nature of the ground truth BEV. This discrepancy imposes constraints on the perception capabilities of the model, leading to inherent limitations. To mitigate this shortfall, we introduce the Ground Truth BEV (GT-BEV) module. The core objective of the GT-BEV is to align the generated BEV representation with the GT-BEV, ensuring an explicit arrangement of BEV elements based on their class label, location, and boundary.

In GT-BEV, we first employ a ground truth encoder $GTEnc$ to represent the class label $c^i$ and ground-truth bounding box $p^i$ information of the $i^{th}$ instance on the BEV map as expressed in Eq.~(\ref{eq:gt-feat}):
\begin{equation}
\beta^{i} = GTEnc(c^i, p^{i}),
\label{eq:gt-feat}
\end{equation}
where $\beta^{i}$, which has the same feature dimension $C$ as the BEV feature, denotes the encoded ground truth feature of the $i^{th}$ instance on the BEV map. Notably, in our experiments (as detailed in Sec. ~\ref{sec:Experiments}), we demonstrate the effectiveness of applying either large language models (LLM) or a simple Multi-Layer Perceptrons (MLP) layer as $GTEnc$ in achieving comparable good results.

Subsequently, for each instance, to enhance the clarity of its boundaries on the BEV map, we crop the area within its ground truth bounding box from the BEV feature map. We apply a pooling operation to the cropped tensor, serving as the representation for the corresponding object $\alpha^{i}$, as shown in Eq.~(\ref{eq:obj-bev-feat}):
\begin{equation}
\alpha^{i} = Pool(Crop(z_{bev}, p^{i})).
\label{eq:obj-bev-feat}
\end{equation}

Finally, to pull the BEV and GT-BEV embeddings closer, we employ the contrastive learning procedure~\cite{clip} to optimize the element relationship and distances inside the BEV feature space, as formulated in Eq.~(\ref{eq:bev-contra}). 
\begin{equation}
\begin{aligned}
\mathcal{M} &= \lambda \cdot \frac{\alpha}{\left \| \alpha \right \|_{2}} \otimes  \frac{\beta}{\left \| \beta \right \|_{2}},\\
\mathcal{L}_{GT-BEV} &= \frac{\mathcal{L}_{CE}(\mathcal{M}, \mathcal{I}) + \mathcal{L}_{CE}(\mathcal{I}, \mathcal{M})}{2}, 
\end{aligned}
\label{eq:bev-contra}
\end{equation}
where $\mathcal{M}$ and $\mathcal{I}$ denote the produced and target similarity matrices between object BEV feature and object ground truth feature, respectively, $\lambda$ is the logit scale learned during contrastive learning, $\otimes$ represents the matrix multiplication, $\mathcal{L}_{CE}$ is the cross entropy loss employed for the optimization of the similarity matrix, and $\mathcal{L}_{GT-BEV}$ is the final loss produced from the GT-BEV module. The detailed principle and explanation of the contrastive learning process is explained in CLIP \cite{clip}. The BEV object relationship is guided by the ground truth similarity, with training loss optimized through the similarity matrix between BEV features and GT features. This involves averaging cross-entropy loss along both the BEV and GT axes, ensuring effective alignment.

By orchestrating a meticulous arrangement of similarities in the feature space, our method provides explicit guidance for BEV generation. This explicit element arrangement, coupled with GT guidance (label, position, and clear boundary), serves to elevate the perceptual capabilities of the model, enabling improved detection and differentiation of various objects on the BEV map.

\subsection{Ground Truth Query Interaction}
\label{ssec:gt-query}

A previously unexplored aspect in prior models lies in understanding how ground truth instances interact with each other on the BEV map. The recovery of ground truth information from empty queries through self-attention and cross-attention mechanisms, as employed in previous models, is inherently limited. This process operates as a black-box exploration, providing supervision solely at the endpoint, thus lacking a comprehensive understanding of ground truth decoding.
To address this limitation, we introduce the Ground Truth Query Interaction (GT-QI) module, a novel addition to our CLIP-BEVFormer framework. The primary objective of the GT-QI module is to inspire the decoder parameter learning process by performing interactions among ground truth instances. 

In the GT-QI module, the encoded ground truth features $\beta$ from the Ground Truth Encoder (GTEnc) are introduced into the query pool of the decoder $Dec$, undergoing the same processes and modules as normal queries. This mimics real information communication in actual scenarios through self-attention (SA) mechanisms within the ground truth queries and global environment communication through cross-attention (CA) with the BEV map. Further processing of the ground truth query information is performed using a Feedforward Neural Network (FFN). It is worth noting that although the procedure and module weights are shared, the attention during decoding is executed in a parallel manner, preventing information leakage. The ground truth flow only influences and inspires the learning of decoder parameters. Since the ground truth query is initially targeted at a specific instance, no matching procedure is required for the ground truth query flow. After decoding, the same head $Head$ and perception loss $\mathcal{L}_{perc}$ are applied to the processed ground truth queries.
The above process can be formulated as Eq.~(\ref{eq:gt-qi}):
\begin{equation}
\begin{aligned}
q_{gt} = \beta, & \quad q_{gt}^{'} = Dec(q_{gt}), \\
q_{gt}^{pred} &= Head(q_{gt}^{'}), \\
\mathcal{L}_{GT-QI} &= \mathcal{L}_{perc}(q_{gt}^{pred}, y),
\end{aligned}
\label{eq:gt-qi}
\end{equation}
where $q_{gt}$ denotes the initial GT queries, $q_{gt}^{'}$ and $q_{gt}^{pred}$ indicate the GT queries after processing of perception decoder and perception head respectively, and $\mathcal{L}_{GT-QI}$ is the final loss generated by GT-QI module.

With ground truth flow injected during the decoding phase of training, our GT-QI module enables the modules to gain insights from both ground truth inter-instance interaction and ground truth instance-BEV communication. The enlarged query pool, injected with GT queries, not only enhances the robustness of the model but also augments its ability for detecting various objects on the source map.

\subsection{Loss}
\label{ssec:loss}
The training loss formulation for our CLIP-BEVFormer encompasses three key components, each designed to optimize specific aspects of the model's performance: baseline BEV detector loss $\mathcal{L}_{base}$, Ground Truth BEV supervision loss $\mathcal{L}_{GT-BEV}$, perception loss of Ground Truth Query Interaction $\mathcal{L}_{GT-QI}$, as expressed in Eq.~(\ref{eq:loss}):
\begin{equation}
\mathcal{L}=\mathcal{L}_{base}+\mathcal{L}_{GT-BEV}+\mathcal{L}_{GT-QI}.
\label{eq:loss}
\end{equation}
The ground truth flow, applied exclusively during the training phase via GT-BEV and GT-QI, introduces no additional parameters or computations during inference. This ensures that the efficiency of the original model is maintained at the inference stage.

\section{Experiments}
\label{sec:Experiments}

\subsection{Implementation Details}
\label{ssec:setting}
\textbf{Dataset.} Our experiments are conducted on public nuScenes dataset\cite{nuscenes}, which is a commonly used dataset in autonomous driving area. It contains 1000 scenes, each with a $\sim$20s duration, and the keyframes are annotated at 2Hz. Each scene is captured with 6 cameras covering the entire 360 $^{\circ}$ field-of-view. Overall, the nuScenes dataset contains 1.4M 3D annotated bounding boxes of 10 object categories. 

\noindent\textbf{Metrics.} We focus our experiments on the 3D object detection task within the nuScenes dataset. This task involves placing a 3D bounding box around objects belonging to 10 distinct categories while estimating various attributes and the current velocity vector. For the evaluation of performance, we employ a set of metrics, including mean Average Precision (mAP), Average Translation Error (ATE), Average Scale Error (ASE), Average Orientation Error (AOE), Average Velocity Error (AVE), Average Attribute Error (AAE), and nuScenes detection score (NDS), as defined in \cite{nuscenes}.

\noindent\textbf{Experiment Set-up.} We apply the typical BEV detector, BEVFormer\cite{bevformer}, as our baseline model. To maintain consistency, we adopt the same training hyperparameters as outlined in \cite{bevformer}. This choice allows for a fair comparison and highlights the improvements brought by our proposed CLIP-BEVFormer framework. For training with language model as GT encoder, we freeze the whole model except the last projection layer, while for training with MLP as GT encoder, we train the whole layer. 
More details regarding the experiment set-up can be found in supplementary materials.

\subsection{3D Detection Results}
\label{ssec:3d-det}

We compare our proposed CLIP-BEVFormer against the baseline BEVFormer and other state-of-the-art BEV detectors listed in Tab.~\ref{tab:3d-val}. We assess the performance across different model configurations, specifically applying CLIP-BEVFormer to both tiny and base variants of BEVFormer. Additionally, we explore the impact of employing either a language model (LM) in the pretrained CLIP \cite{clip} or a simple MLP layer as the ground truth (GT) encoder.

As illustrated in the Tab.~\ref{tab:3d-val}, all four variants of CLIP-BEVFormer consistently outperform the baseline models accross all metrics except a slight degradation for tiny configuration in mAAE metric. The improvements are more evident in both nuScenes detection scores (NDS) and mean Average Precision (mAP). Specificially for tiny configuration, in comparison with baseline BEVFormer, CLIP-BEVFormer with MLP achieves 8.5\% and 9.2\% improvement in terms of NDS and mAP, respectively.  Similarly, CLIP-BEVFormer with LM demonstrates 9.3\% and 8.8\% improvement in NDS and mAP, respectively, compared to counterpart BEVFormer. For base configuration, CLIP-BEVformer shows similar consistent improvement over the baseline, with our MLP variant showing best results. 

The observed enhancements across diverse model configurations emphasize the effectiveness of injecting GT-flow for inspiring both BEV encoding and perception decoding processes in BEV detectors.
The comparable performances achieved by both MLP and LM variants indicate that our framework is robust and less sensitive to the choice of the ground truth encoder. This flexibility makes CLIP-BEVFormer adaptable and easily deployable for integration with various detectors.
The consistent improvements across all variants signify the robustness of CLIP-BEVFormer, demonstrating its capability to enhance 3D object detection tasks under different model complexities and encoder choices.

\begin{table*}[t]
\centering
\resizebox{1.0\linewidth}{!}{
    \begin{tabular}{l|ccc|cc|ccccc}
        \toprule
        Method & GT Enc & Modality & Backbone & NDS $\uparrow$ & mAP$\uparrow$ & mATE$\downarrow$ & mASE$\downarrow$ & mAOE$\downarrow$ & mAVE$\downarrow$ & mAAE$\downarrow$ \\
        \midrule
        SSN~\cite{55_zhu2020ssn} & - & Lidar & PointPollar & 49.8 & 36.6 & - & - & - & - & - \\
        CenterPoint-Voxel~\cite{52_yin2021center} & - & Lidar & VoxelNet & 64.8 & 56.4 & - & - & - & - & - \\
        \midrule
        FCOS3D~\cite{45_wang2021fcos3d} & - & Cam & R101 & 41.5 & 34.3 & 0.725 & 0.263 & 0.422 & 1.292 &0.153 \\
        PGD~\cite{44_wang2022probabilistic} & - & Cam & R101 & 42.8 & 36.9 & 0.683 & 0.260 & 0.439 & 1.268 &0.185 \\
        DETR3D~\cite{47_wang2022detr3d} & - & Cam & R101 & 42.5 & 34.6 & 0.773 & 0.268 & 0.383 & 0.842 &0.216 \\
        \midrule
        BEVerse \cite{beverse} & - & Cam & Swin & 46.6 & 32.1 & 0.681 & 0.278 & 0.466 & 0.328 & 0.190 \\
        \rowcolor{blue!15} +Ours & LM & Cam & Swin & \textbf{48.3} & \textbf{34.2} & \textbf{0.665} & \textbf{0.270} & \textbf{0.456} & \textbf{0.318} & \textbf{0.170} \\
        \midrule
        BEVFormer-tiny\cite{bevformer} & - & Cam & R50 & 35.5 & 25.1 & 0.898 & 0.293 & 0.651 & 0.657 & \underline{0.216} \\
        \rowcolor{blue!15} +Ours & MLP & Cam & R50 & \textbf{38.5} & \underline{\textbf{27.4}} & \textbf{0.869} & \textbf{0.283} & \textbf{0.607} & \textbf{0.542} & 0.220 \\
        \rowcolor{blue!15} +Ours & LM & Cam & R50 & \underline{\textbf{38.8}} & \textbf{27.3} & \underline{\textbf{0.856}} & \underline{\textbf{0.282}} & \underline{\textbf{0.583}} & \underline{\textbf{0.538}} & 0.228 \\
        \midrule
        BEVFormer-base\cite{bevformer} & - & Cam & R101 & 51.7 & 41.6 & 0.673 & 0.274 & 0.372 & 0.394 & 0.198 \\
        \rowcolor{blue!15} +Ours & MLP & Cam & R101 & \underline{\textbf{56.2}} & \underline{\textbf{46.7}} & \underline{\textbf{0.605}} & \underline{\textbf{0.253}} & \textbf{0.331} & \textbf{0.336} & \textbf{0.187} \\
        \rowcolor{blue!15} +Ours & LM & Cam & R101 & \textbf{55.1} & \textbf{44.1} & \textbf{0.641} & \underline{\textbf{0.253}} & \underline{\textbf{0.319}} & \underline{\textbf{0.307}} & \underline{\textbf{0.172}} \\
        \midrule
        BEVformerV2 \cite{bevformerv2} & - & Cam & R50 & 42.6 & 35.1 & 0.753 & 0.286 & 0.466 & 0.807 & \textbf{0.186} \\
        \rowcolor{blue!15} +Ours & LM & Cam & R50 & \textbf{44.1} & \textbf{37.0} & \textbf{0.729} & \textbf{0.281} & \textbf{0.438} & \textbf{0.791} & 0.204 \\
        \bottomrule
    \end{tabular}
}
\caption{\textbf{3D Detection results on nuScenes validation set.} Comparison of 3D object detection performance across various state-of-the-art BEV detectors. CLIP-BEVFormer, applied to both tiny and base variants of BEVFormer, demonstrates consistent improvements over baseline models. Results are reported for different ground truth encoder choices, including Language Model (LM) and Multi-Layer Perceptron (MLP). Similar improvements achieved through LM and MLP encoders underscores the generalizability of proposed framework.  }
\label{tab:3d-val}
\end{table*}

\subsection{Long-tail Detection Results}
\label{ssec:long-tail}

To assess the effectiveness and generalization ability of our proposed method in handling long-tail cases, we present per-class detection results in Tab.~\ref{tab:long-tail}. The nuScenes dataset exhibits a considerable class imbalance, where some classes represent a small portion (1\%) and others a large portion (43\%) of the dataset, as detailed in the Tab.~\ref{tab:long-tail}. We evaluate the performance of CLIP-BEVFormer across specific object classes to highlight its ability to address challenges associated with less common occurrences.

As shown in the Tab.~\ref{tab:long-tail}, CLIP-BEVFormer demonstrates an overall improvement in 3D detection performance across all classes, which indicates that our method positively contributes to the perception ability of the model, showcasing its robustness and adaptability, irrespective of the class distribution.
Notably, for long-tail classes such as construction vehicle, bus, motorcycle, bicycle, and trailer, in which each of these categories account for approximately 1\% of the dataset, CLIP-BEVFormer demonstrates substantial improvements. In particular, our CLIP-BEVFormer-base with simple MLP as ground truth encoder improves detection performance  of construction vehicle, bus, motorcycle, bicycle, and trailer classes by 46.5\%, 14.4\%, 15.6\%, 10.5\%, and 26.7\%, respectively, in comparison with the baseline BEVformer.

The considerable enhancements observed in long-tail classes underscore the enhanced learning ability and reduced sensitivity to data imbalance of CLIP-BEVFormer. By addressing challenges associated with less common object occurrences, our method showcases its potential for real-world deployment, where imbalanced class distributions are common. The ability to improve detection accuracy in long-tail scenarios further establishes CLIP-BEVFormer as a robust and reliable solution for 3D object detection tasks in autonomous driving systems.

\begin{table*}[t]
\centering
\resizebox{1.0\linewidth}{!}{
    \begin{tabular}{l|c|cccccccccc}
        \toprule
          & GT Enc & CV & BUS & MOT & BIC & TR & TUK & CONE & BAR & PED & CAR \\
        \midrule
        Total Num & & 650 & 657 & 748 & 857 & 1114 & 4215 & 6591 & 10263 & 11564 & 27727 \\
        Percentage & & 1.0\% & 1.0\% & 1.2\% & 1.3\% & 1.7\% & 6.5\% & 10.2\% & 15.9\% & 18.0\% & 43.1\% \\
        \midrule
        BEVFormer-tiny~\cite{bevformer} & - & 5.8 & 23.3 & 21.4 & 20.3 & 6.6 & 19.2 & 38.4 & 37.9 & 33.2 & 45.7 \\
        \rowcolor{blue!15} CLIP-BEVFormer-tiny & MLP & \underline{\textbf{7.1}} & \textbf{28.0} & \underline{\textbf{26.1}} & \textbf{21.6} & \textbf{8.1} & \textbf{20.9} & \underline{\textbf{41.1}} & \underline{\textbf{40.0}} & \textbf{33.9} & \underline{\textbf{46.8}}  \\
        \rowcolor{blue!15} CLIP-BEVFormer-tiny & LM &  \textbf{6.0} & \underline{\textbf{28.3}} & \textbf{25.2} & \underline{\textbf{22.2}} & \underline{\textbf{8.7}} & \underline{\textbf{22.2}} & \textbf{40.5} & \textbf{39.3} & \underline{\textbf{34.3}}& \textbf{46.6} \\
        \midrule
        BEVFormer-base~\cite{bevformer} & - & 12.9 & 44.4 & 42.9 & 39.8 & 17.2 & 37.0 & 58.4 & 52.5 & 49.4 & 61.8\\
        \rowcolor{blue!15} CLIP-BEVFormer-base & MLP & \underline{\textbf{18.9}} & \underline{\textbf{50.8}} & \underline{\textbf{49.6}} & \underline{\textbf{44.0}} & \underline{\textbf{21.8}} & \underline{\textbf{40.9}} & \underline{\textbf{63.1}} & \textbf{55.7} & \underline{\textbf{55.2}} & \underline{\textbf{66.6}} \\
        \rowcolor{blue!15} CLIP-BEVFormer-base & LM & \textbf{14.0} & \textbf{46.9} & \textbf{46.6} & \textbf{41.1} & \textbf{19.6} & \textbf{37.9} & \textbf{62.6} & \underline{\textbf{56.4}} & \textbf{52.1} & \textbf{64.6} \\
        \bottomrule
    \end{tabular}
}
\caption{\textbf{Per-class 3D detection results on NuScenes validation set.} Evaluation of per-class 3D object detection results, showcasing the performance of CLIP-BEVFormer and baseline BEVFormer on long-tail cases. The table provides insights into the distribution, total numbers, and percentages of instances across specific object classes. CLIP-BEVFormer exhibits overall enhancements, particularly more pronounced improvement in classes with lower($\sim$1-2\%) occurrence frequencies, highlighting its efficacy in addressing imbalanced class distributions. The CV, BUS, MOT, BIC, TR, TUK, CONE, BAR, PED, and CAR denote the construction vehicle, bus, motorcycle, bicycle, trailer, truck, traffic cone, barrier, pedestrian, and car, respectively.}
\label{tab:long-tail}
\end{table*}

\begin{figure*}[bt!]
  \centering\includegraphics[width=1.0\linewidth]{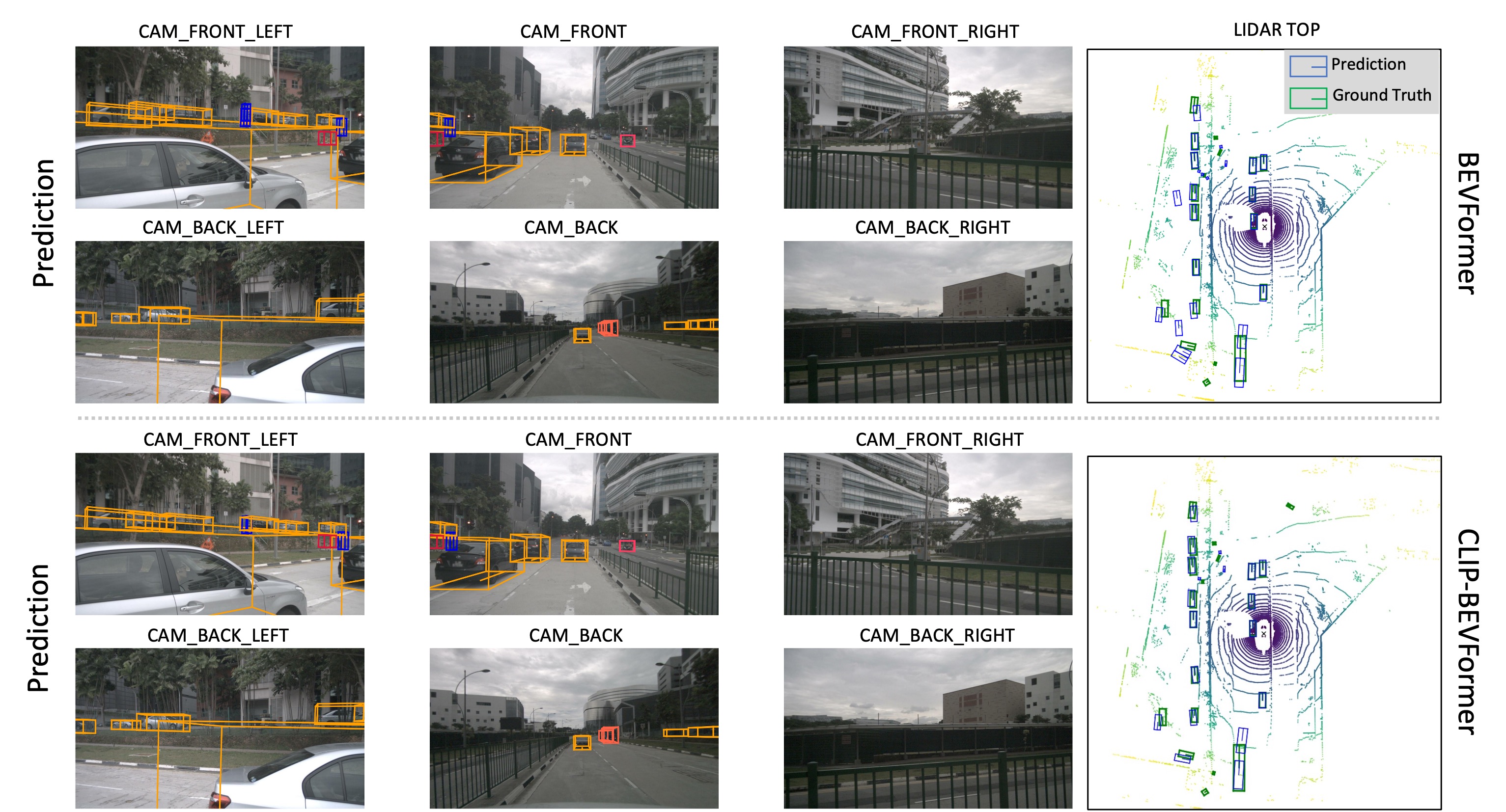}
   \caption{\textbf{Visualization results on nuScenes validation set.} We demonstrate qualitative detection performance on both camera and BEV images. As can be seen in BEV images, CLIP-BEVFormer demonstrates improved alignment with ground truth detections.}
   \label{fig:vis}
\end{figure*}

\subsection{Robustness Results}
\label{ssec:robust}

In real-world deployment scenarios, autonomous driving systems must contend with potential sensor failures arising from hardware malfunctions, adverse weather conditions, or physical obstructions. Evaluating the detector's performance under such conditions is crucial for assessing its robustness. Thus, we design and conduct first sensor failure robustness study. To simulate sensor failures, we conducted experiments involving the random masking of one camera view during inference, mimicking scenarios where a camera might malfunction.

The robustness evaluation metrics include NDS, mAP, mATE, mASE, mAOE, mAVE, mAAE, and the mAP on the long-tail classes. As depicted in Tab.~\ref{tab:robustness}, CLIP-BEVFormer consistently outperforms the baseline models in various configurations. Specifically, for tiny configuration, CLIP-BEVFormer with the language model (LM) as the ground truth encoder achieves 12.3\% NDS, 15.7\% mAP, and 27.5\% long-tail mAP  improvements over the baseline BEVFormer, demonstrating its superior performance and enhanced robustness under the simulated sensor failure scenario.

\begin{table*}[t]
\centering
\resizebox{1.0\linewidth}{!}{
    \begin{tabular}{l|c|cc|ccccc|c}
        \toprule
        Method & GT Enc & NDS$\uparrow$ & mAP$\uparrow$ & mATE$\downarrow$ & mASE$\downarrow$ & mAOE$\downarrow$ & mAVE$\downarrow$ & mAAE$\downarrow$ & Long-Tail$\uparrow$\\
        \midrule
        BEVFormer-tiny~\cite{bevformer} & - & 33.3 & 21.7 & 0.908 & 0.295 & 0.656 & 0.681 & \underline{0.218} & 13.1 \\
        \rowcolor{blue!15} CLIP-BEVFormer-tiny & MLP & \textbf{36.6} & \textbf{24.3} & \textbf{0.881} & \textbf{0.284} & \textbf{0.617} & \textbf{0.550} & 0.225 & \textbf{16.2} \\
        \rowcolor{blue!15} CLIP-BEVFormer-tiny & LM & \underline{\textbf{37.4}} & \underline{\textbf{25.1}} & \underline{\textbf{0.873}} & \underline{\textbf{0.283}} & \underline{\textbf{0.589}} & \underline{\textbf{0.540}} & 0.230 & \underline{\textbf{16.7}} \\
        \midrule
        BEVFormer-base~\cite{bevformer} & - & 49.6 & 38.0 & 0.686 & 0.275 & 0.379 & 0.395 & 0.201 & 28.3 \\
        \rowcolor{blue!15} CLIP-BEVFormer-base & MLP & \underline{\textbf{53.4}} & \underline{\textbf{42.1}} & \underline{\textbf{0.612}} & \textbf{0.273} & \underline{\textbf{0.348}} & \underline{\textbf{0.341}} & \textbf{0.192} & \underline{\textbf{31.7}} \\
        \rowcolor{blue!15} CLIP-BEVFormer-base & LM & \textbf{52.0} & \textbf{40.2} & \textbf{0.663} & \underline{\textbf{0.263}} & \textbf{0.355} & \underline{\textbf{0.341}} & \underline{\textbf{0.183}} & \textbf{30.5} \\
        \bottomrule
    \end{tabular}
}
\caption{\textbf{Robustness evaluation results.} The robustness study experiments is conducted by random masking of one camera view during inference to simulate practical deployment scenarios with potential sensor failures. Proposed CLIP-BEVFormer shows  improved robustness performance over the BEVFormer baseline for both tiny and base configurations.}
\label{tab:robustness}
\end{table*}

\subsection{Ablation Study}
\label{ssec:abl}

\textbf{GT encoder and GT input format.}
We investigate the ground truth input format along with the ground truth encoder.
Three distinct input formats, namely digit, semantic, and scene, are explored. For the digit format, we directly utilize ground truth digit numbers, encompassing the one-hot format of the class label and the normalized 3D bounding box for each instance. In the semantic format, a semantic sentence template, such as ``This is a \{class-label\}. Its 3D bounding box is \{3D-bbox location\}," is employed as the GT input. For the scene format, we augment the semantic input with a scene description provided in the dataset, i.e. ``This is a \{class-label\}. Its 3D bounding box is \{3D-bbox location\}. Its scene description is \{scene\}.". 
For the ground truth encoder, we investigate different configurations, including the language model in pretrained CLIP~\cite{clip}, GPT-2~\cite{gpt2}, and a simple MLP layer. 

As depicted in Tab.~\ref{tab:abl-prompt}, all configurations outperform the baselines, demonstrating that GT flow provides crucial guidance to the detector. Notably, the results reveal that this guidance is less sensitive to the GT encoder and the GT input format. Regardless of the specific configuration, the performance remains consistently improved across both tiny and base variants, emphasizing the robust and flexible nature of our proposed CLIP-BEVFormer.

\begin{table}[t]
\centering
\resizebox{1.0\linewidth}{!}{
    \begin{tabular}{l|c|cc|cc}
        \toprule
        \multirow{2}{*}{GT Enc} & \multirow{2}{*}{GT Format} & \multicolumn{2}{c|}{Tiny} & \multicolumn{2}{c}{Base} \\
         & & NDS$\uparrow$ & mAP$\uparrow$ & NDS$\uparrow$ & mAP$\uparrow$ \\
        \midrule
        - & - & 35.5 & 25.1 & 51.7 & 41.6 \\
        \midrule
        \rowcolor{blue!15} MLP & digit & \textbf{38.5} & \textbf{27.4} & \underline{\textbf{56.2}} & \underline{\textbf{46.7}} \\
        \rowcolor{blue!15} CLIP-LM & digit & \textbf{38.4} & \textbf{27.6} & \textbf{55.9} & \textbf{45.3} \\
        \rowcolor{blue!15} CLIP-LM & semantic & \underline{\textbf{38.8}} & \textbf{27.3} & \textbf{55.1} & \textbf{44.1} \\
        \rowcolor{blue!15} GPT2 & semantic & \textbf{37.1} & \textbf{26.3} & \textbf{54.8} & \textbf{45.1} \\
        \rowcolor{blue!15} CLIP-LM & scene & \textbf{38.2} & \underline{\textbf{27.7}} & \textbf{55.3} & \textbf{45.0} \\
        \bottomrule
    \end{tabular}
}
\caption{\textbf{Ablation study on ground truth encoder and ground truth input format.} CLIP-BEVFormer consistently shows improvement over the baseline across variety of ground truth encoders and input formats.}
\label{tab:abl-prompt}
\vspace{-0.3cm}
\end{table}

\noindent\textbf{Effectiveness of each GT guidance.}
To assess the impact of each Ground Truth (GT) guidance component in CLIP-BEVFormer, we conduct ablation studies on both the tiny and base variants, as presented in Tab. ~\ref{tab:abl-model-config}. The GT encoder configurations include a Multilayer Perceptron (MLP) and a Language Model (LM). The ablation study explores the effectiveness of GT guidance solely for BEV encoding (BEV) and both BEV encoding and perception decoding (BEV \& Dec).
Results indicate that incorporating only GT-BEV guidance yields notable improvements over the baseline, enhancing NDS by 5.9\% and 5.0\%, and mAP by 7.2\% and 2.9\%, with LM as GT encoder, for both tiny and base models, respectively. Further enhancement is observed when integrating both GT guidance components, resulting in superior performance. This consistency across various configurations emphasizes the robust and effective nature of our proposed GT guidance in enhancing both BEV encoding and perception decoding processes.

\begin{table}[t]
\centering
\resizebox{1.0\linewidth}{!}{
    \begin{tabular}{l|c|cc|cc}
        \toprule
        \multirow{2}{*}{GT Enc} & \multirow{2}{*}{GT Guidance} & \multicolumn{2}{c|}{Tiny} & \multicolumn{2}{c}{Base} \\
         & & NDS$\uparrow$ & mAP$\uparrow$ & NDS$\uparrow$ & mAP$\uparrow$ \\
        \midrule
        - & - & 35.5 & 25.1 & 51.7 & 41.6 \\
        \midrule
        \rowcolor{blue!15} MLP & BEV & \textbf{37.3} & \textbf{26.9} & \textbf{55.2} & \textbf{43.2} \\
        \rowcolor{blue!15} LM & BEV & \textbf{37.6} & \textbf{26.9} & \textbf{54.3} & \textbf{42.8} \\
        \rowcolor{blue!15} MLP & BEV \& Dec & \textbf{38.5} & \underline{\textbf{27.4}} & \underline{\textbf{56.2}} & \underline{\textbf{46.7}}\\
        \rowcolor{blue!15} LM & BEV \& Dec & \underline{\textbf{38.8}} & \textbf{27.3} & \textbf{55.1} & \textbf{44.1} \\
        \bottomrule
    \end{tabular}
}
\caption{\textbf{Ablation study for effectiveness of GT-BEV and GT-QI.} Results show the importance of both ground truth guidance components in achieving improved detection accuracy. }
\label{tab:abl-model-config}
\vspace{-0.3cm}
\end{table}

\subsection{Qualitative Results}
\label{ssec:vis}

The visual comparative results presented in Fig.~\ref{fig:vis} illustrates the enhanced performance of CLIP-BEVFormer in generating better Bird's Eye View (BEV) representations compared to the baseline BEVFormer. The visualized detection results shows that the output of CLIP-BEVFormer aligns more closely with the Ground Truth BEV, indicating an improvement in the precision of BEV generation. The qualitative results affirm that our method effectively enhances both the quality of BEV representation and the accuracy of 3D object detection.

\section{Conclusion and Future Work}
\label{sec:Conclusion}
\textbf{Conclusion:} 
In this study, we introduce CLIP-BEVFormer, a pioneering framework aimed at advancing multi-view image-based Bird's Eye View (BEV) detectors. Comprising the GT-BEV module and GT-QI module, CLIP-BEVFormer leverages ground truth flow guidance in both the BEV encoding and perception decoding processes. The GT-BEV module orchestrates explicit arrangements for BEV elements, driven by class labels, locations, and boundaries, effectively elevating the BEV to approximate ground truth structures. Simultaneously, the GT-QI module enriches the decoder query pool and inspires the perception learning process. Our extensive experiments on the challenging nuScenes dataset demonstrate the consistent superiority of CLIP-BEVFormer over various tasks.

\noindent\textbf{Limitations and Future Work:} 
The current study establishes a strong foundation for enhancing BEV detectors through CLIP-BEVFormer. In the future work, we will investigate its application to different sensor modalities such as LiDAR.
Additionally, the framework will be generalized to encompass a broader spectrum of autonomous driving tasks, including object tracking, scene segmentation, and motion prediction. Continuous refinement and adaptation will be pursued to uphold the robustness and versatility of CLIP-BEVFormer in real-world scenarios.

\clearpage
\setcounter{page}{1}
\maketitlesupplementary

\section{Experiment Setup}
\label{sec:setting}

In our experiments, we adopt ResNet50 and ResNet101 \cite{resnet} as backbones for the BEVFormer-tiny and BEVFormer-base models, respectively. These backbones are initialized from the FCOS3D \cite{fcos3d} checkpoint, following the configuration in BEVFormer \cite{bevformer}. We leverage the output multi-scale features from the Feature Pyramid Network (FPN) \cite{fpn}, with sizes of 1/16, 1/32, and 1/64, and the dimension of 256. During the training phase for CLIP-BEVFormer, we leverage the pretrained language model in CLIP-RN101~\cite{clip} as our off-the-shelf language model.

The BEV size for the tiny and base variants is set to $50 \times 50$ and $200 \times 200$, respectively, while the perception ranges span from -51.2m to 51.2m along the X and Y axes. The resolution of the BEV grid is set to 0.512m. We incorporate learnable positional embeddings for BEV queries to enrich the spatial representation.

The BEV encoder comprises 6 encoder layers, consistently refining BEV queries in each layer. During the spatial cross-attention module, implemented using the deformable attention mechanism, each local query corresponds to four target points with different heights in 3D space. The predefined height anchors are uniformly sampled from -5 meters to 3 meters.

For each reference point on 2D view features, we utilize four sampling points around this reference point for each head. During training, we use a 2-frame history BEV for the tiny variant and a 3-frame history BEV for the base variant. We train our models for 24 epochs with a learning rate of $2 \times 10^{-4}$ \cite{bevformer}.

\section{3D Object Detection Results with Various Baselines}
\label{sec:3d-det-test}

We have conducted experiments with various detection baselines, BEVformer \cite{bevformer}, BEVformerV2 \cite{bevformerv2}, and BEVerse \cite{beverse}. We evaluate our model on both validation and test sets of nuScenes. The results presented in Tab.~\ref{tab:more-det-baselines} show that our method consistently improves the perception capabilities of various baselines by significant margins on both sets, indicating its flexibility and model-agnostic nature. 

\begin{table*}[hb]
\centering
\resizebox{1.0\linewidth}{!}{
    \begin{tabular}{l|c|cc|ccccc|cc|ccccc}
        \toprule
        \multirow{2}{*}{Model} & \multirow{2}{*}{Backbone} & \multicolumn{7}{c|}{Validation Set} & \multicolumn{7}{c}{Test Set} \\
         &  & NDS $\uparrow$ & mAP$\uparrow$ & mATE$\downarrow$ & mASE$\downarrow$ & mAOE$\downarrow$ & mAVE$\downarrow$ & mAAE$\downarrow$ & NDS $\uparrow$ & mAP$\uparrow$ & mATE$\downarrow$ & mASE$\downarrow$ & mAOE$\downarrow$ & mAVE$\downarrow$ & mAAE$\downarrow$ \\
        \midrule
BEVformer-tiny & R50 & 35.5 & 25.1 & 0.898 & 0.293 & 0.651 & 0.657 & \textbf{0.216} & 37.2 & 27.3 & 0.856 & 0.283 & 0.609 & 0.753 & 0.146 \\
\rowcolor{blue!15} +Ours & R50 & \textbf{38.8} & \textbf{27.3} & \textbf{0.856} & \textbf{0.282} & \textbf{0.583} & \textbf{0.538} & 0.228 & \textbf{41.1} & \textbf{29.3} & \textbf{0.811} & \textbf{0.271} & \textbf{0.554} & \textbf{0.579} & \textbf{0.136} \\
\midrule
BEVformer-base & R101 & 51.7 & 41.6 & 0.673 & 0.274 & 0.372 & 0.394 & 0.198 & 53.5 & 44.5 & 0.631 & 0.257 & \textbf{0.405} & 0.435 & 0.143 \\
\rowcolor{blue!15} +Ours & R101 & \textbf{55.1} & \textbf{44.1} & \textbf{0.641} & \textbf{0.253} & \textbf{0.319} & \textbf{0.307} & \textbf{0.172} & \textbf{54.7} & \textbf{44.7} & \textbf{0.591} & \textbf{0.257} & 0.417 & \textbf{0.371} & \textbf{0.128} \\
\midrule
BEVformerV2 & R50 & 42.6 & 35.1 & 0.753 & 0.286 & 0.466 & 0.807 & \textbf{0.186} & 42.5 & 35.4 & 0.707 & 0.278 & 0.506 & \textbf{0.895} & \textbf{0.134} \\
\rowcolor{blue!15} +Ours & R50 & \textbf{44.1} & \textbf{37.0} & \textbf{0.729} & \textbf{0.281} & \textbf{0.438} & \textbf{0.791} & 0.204 & \textbf{43.6} & \textbf{37.9} & \textbf{0.676} & \textbf{0.272} & \textbf{0.475} & 0.975 & 0.141 \\
\midrule
BEVerse & Swin & 46.6 & 32.1 & 0.681 & 0.278 & 0.466 & 0.328 & 0.190 & 50.1 & 36.2 & 0.610 & 0.257 & 0.451 & 0.355 & 0.131 \\
\rowcolor{blue!15} +Ours & Swin & \textbf{48.3} & \textbf{34.2} & \textbf{0.665} & \textbf{0.270} & \textbf{0.456} & \textbf{0.318} & \textbf{0.170} & \textbf{52.2} & \textbf{37.4} & \textbf{0.556} & \textbf{0.247} & \textbf{0.413} & \textbf{0.301} & \textbf{0.129} \\
        \bottomrule
    \end{tabular}
}
\caption{3D object detection results on nuScenes validation and test sets.}
\label{tab:more-det-baselines}
\end{table*}

\section{3D Object Detection Metrics}
We adhere to standard evaluation metrics for 3D detection on the nuScenes dataset \cite{nuscenes}, encompassing metrics such as mean Average Precision (mAP), Average Translation Error (ATE), Average Scale Error (ASE), Average Orientation Error (AOE), Average Velocity Error (AVE), Average Attribute Error (AAE), and nuScenes detection score (NDS).

\noindent\textbf{Mean Average Precision (mAP).} For mAP, we utilize the Average Precision metric, modifying the definition of a match by considering the 2D center distance on the ground plane instead of intersection over union-based affinities. Specifically, we match predictions with ground truth objects based on the smallest center distance within a certain threshold. Average precision (AP) is calculated by integrating the recall vs precision curve for recalls and precisions $>$ 0.1. We then average over match thresholds of {0.5, 1, 2, 4} meters and compute the mean across classes.

\noindent\textbf{True Positives (TP).} TP metrics are designed to measure translation, scale, orientation, velocity, and attribute errors. These are calculated using a threshold of 2m center distance during matching and are positive scalars. Metrics are defined per class, and we then take the mean over classes to calculate mATE, mASE, mAOE, mAVE, and mAAE.

\begin{itemize}
\item \textbf{Average Translation Error (ATE).} Euclidean center distance in 2D in meters.
\item \textbf{Average Scale Error (ASE).} Calculated as 1 - IOU after aligning centers and orientation.
\item \textbf{Average Orientation Error (AOE).} Smallest yaw angle difference between prediction and ground truth in radians. Orientation error is evaluated at 360 degrees for most classes, except barriers, where it is evaluated at 180 degrees. Orientation errors for cones are ignored.
\item \textbf{Average Velocity Error (AVE).} Absolute velocity error in m/s. Velocity error for barriers and cones is ignored.
\item \textbf{Average Attribute Error (AAE).} Calculated as 1 - acc, where acc is the attribute classification accuracy. Attribute error for barriers and cones is ignored.
\end{itemize}

\noindent\textbf{nuScenes Detection Score (NDS).} We consolidate the above metrics by computing a weighted sum: mAP, mATE, mASE, mAOE, mAVE, and mAAE. As a first step, we convert TP errors to TP scores using TP\_score = max(1 - TP\_error, 0.0). We then assign a weight of 5 to mAP and 1 to each of the 5 TP scores, calculating the normalized sum.

\section{Training and Inference Efficiency}
Our model is trained with 4 A100 80GB GPUs. Our method does not introduce any additional parameters and computations during the inference stage, which means that it allows for enhanced performance without sacrificing real-time processing capabilities. We provide detailed information on memory, training time, number of parameters and FPS in Tab.~\ref{tab:efficiency}.

\begin{table}[t]
\centering
\resizebox{1.0\linewidth}{!}{
     \begin{tabular}{l|cccc}
        \toprule
        Model & Train Mem (GB) & Train Hrs & \# Params (M) & FPS \\
        \midrule
        BEVformer-tiny & $\sim$7 & $\sim$46 & 33 & 5.1 \\
        +Ours & $\sim$7 & $\sim$46 & 33 & 5.1 \\
        \midrule
        BEVformer-base & $\sim$25 & $\sim$90 & 69 & 2.1 \\
        +Ours & $\sim$25 & $\sim$90 & 69 & 2.1 \\
        \midrule
        BEVformerV2 & $\sim$46 & $\sim$38 & 56 & 2.3 \\
        +Ours & $\sim$46 & $\sim$38 & 56 & 2.3 \\
        \midrule
        BEVerse & $\sim$48 & $\sim$72 & 102.5 & 4.4 \\
        +Ours & $\sim$48 & $\sim$72 & 102.5 & 4.4 \\
        \bottomrule
    \end{tabular}
    }
      \caption{Efficiency details. FPS is tested on 1 V100 GPU.} \label{tab:efficiency}
\end{table}

\section{Visualization}
\label{sec:vis}

\begin{figure*}[bt!]
  \centering\includegraphics[width=1.0\linewidth]{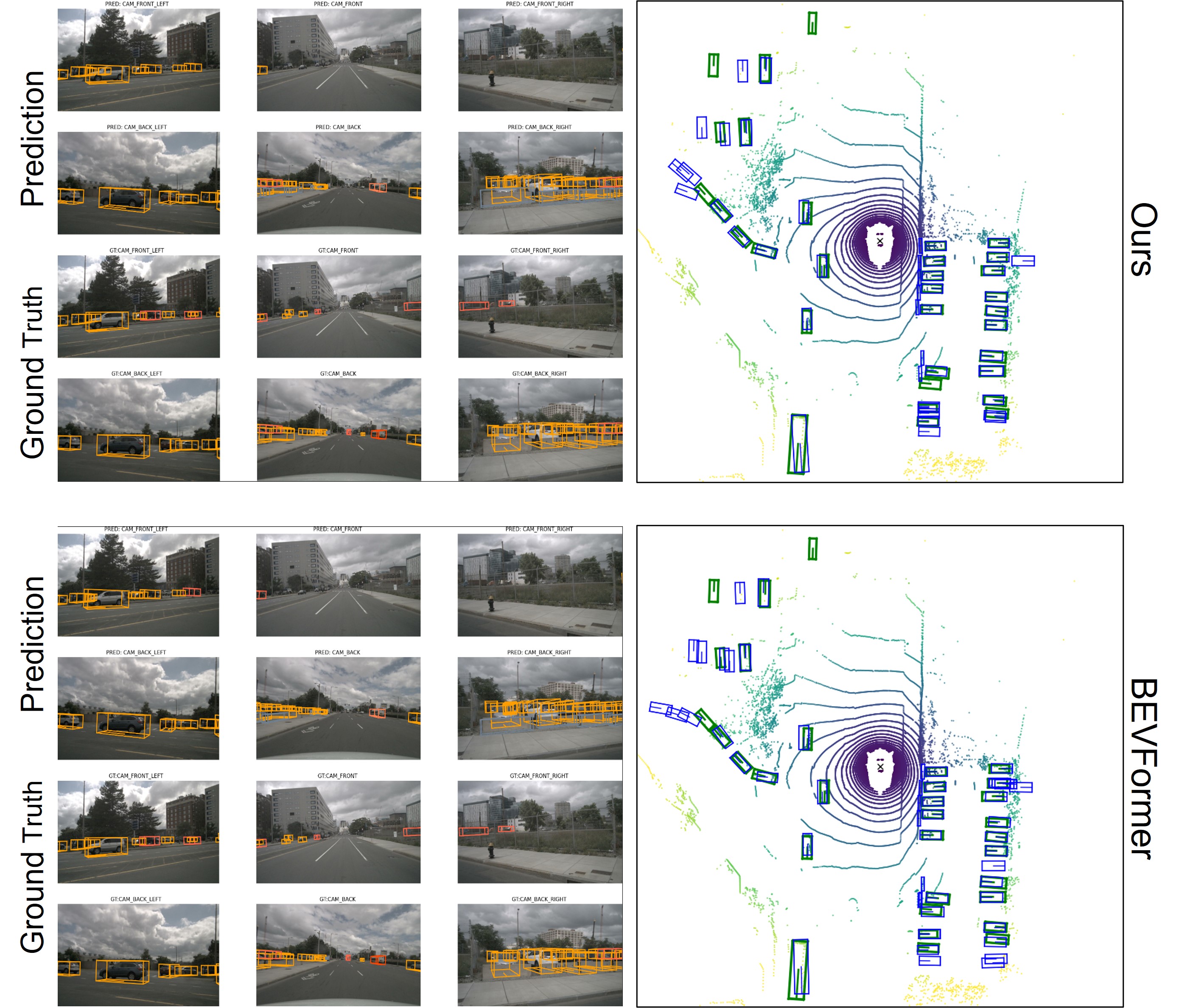}
   \caption{\textbf{Visualization results on nuScenes validation set.} We demonstrate qualitative detection performance on both camera and BEV images. As can be seen in BEV images,  our CLIP-BEVFormer method demonstrates improved alignment with ground truth detections.}
   \label{fig:sup-vis-1}
\end{figure*}

\begin{figure*}[bt!]
  \centering\includegraphics[width=1.0\linewidth]{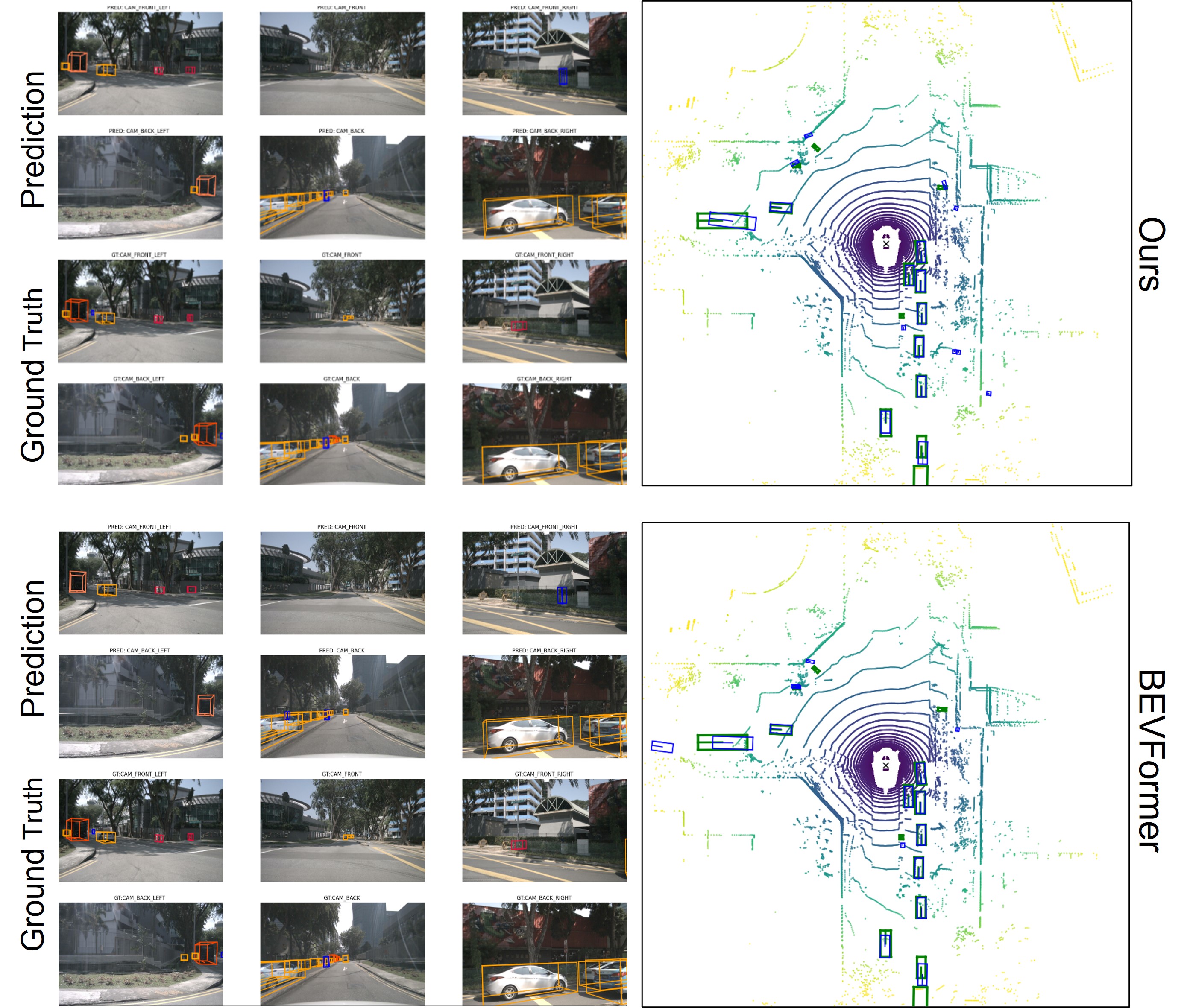}
   \caption{\textbf{Visualization results on nuScenes validation set.} Our CLIP-BEVFormer demonstrates improved alignment with ground truth detections on both camera and BEV images.}
   \label{fig:sup-vis-2}
\end{figure*}

In Fig.~\ref{fig:sup-vis-1} and Fig.~\ref{fig:sup-vis-2}, we present a comprehensive visualization of the qualitative detection performance achieved by CLIP-BEVFormer. The images provide insights into both camera and Bird's Eye View (BEV) perspectives, offering a nuanced understanding of the model's predictions. Notably, these visualizations highlight the enhanced alignment between CLIP-BEVFormer's predictions and ground truth detections in both camera and BEV views, underscoring the model's proficiency in accurately capturing the 3D environment.

{
    \small
    \bibliographystyle{ieeenat_fullname}
    \bibliography{main}

\begin{thebibliography}{46}
\providecommand{\natexlab}[1]{#1}
\providecommand{\url}[1]{\texttt{#1}}
\expandafter\ifx\csname urlstyle\endcsname\relax
  \providecommand{\doi}[1]{doi: #1}\else
  \providecommand{\doi}{doi: \begingroup \urlstyle{rm}\Url}\fi

\bibitem[Alayrac et~al.(2022)Alayrac, Donahue, Luc, Miech, Barr, Hasson, Lenc, Mensch, Millican, Reynolds, et~al.]{flamingo}
Jean-Baptiste Alayrac, Jeff Donahue, Pauline Luc, Antoine Miech, Iain Barr, Yana Hasson, Karel Lenc, Arthur Mensch, Katherine Millican, Malcolm Reynolds, et~al.
\newblock Flamingo: a visual language model for few-shot learning.
\newblock \emph{Advances in Neural Information Processing Systems}, 35:\penalty0 23716--23736, 2022.

\bibitem[Anonymous(2023)]{anonymous2023bevclip}
Anonymous.
\newblock {BEV}-{CLIP}: Multi-modal {BEV} retrieval methodology for complex scene in autonomous driving.
\newblock In \emph{Submitted to The Twelfth International Conference on Learning Representations}, 2023.
\newblock under review.

\bibitem[Caesar et~al.(2020)Caesar, Bankiti, Lang, Vora, Liong, Xu, Krishnan, Pan, Baldan, and Beijbom]{nuscenes}
Holger Caesar, Varun Bankiti, Alex~H Lang, Sourabh Vora, Venice~Erin Liong, Qiang Xu, Anush Krishnan, Yu Pan, Giancarlo Baldan, and Oscar Beijbom.
\newblock nuscenes: A multimodal dataset for autonomous driving.
\newblock In \emph{Proceedings of the IEEE/CVF conference on computer vision and pattern recognition}, pages 11621--11631, 2020.

\bibitem[Carion et~al.(2020)Carion, Massa, Synnaeve, Usunier, Kirillov, and Zagoruyko]{detr}
Nicolas Carion, Francisco Massa, Gabriel Synnaeve, Nicolas Usunier, Alexander Kirillov, and Sergey Zagoruyko.
\newblock End-to-end object detection with transformers.
\newblock In \emph{European conference on computer vision}, pages 213--229. Springer, 2020.

\bibitem[Chen et~al.(2022)Chen, Sima, Li, Zheng, Xu, Geng, Li, He, Shi, Qiao, et~al.]{chen2022persformer}
Li Chen, Chonghao Sima, Yang Li, Zehan Zheng, Jiajie Xu, Xiangwei Geng, Hongyang Li, Conghui He, Jianping Shi, Yu Qiao, et~al.
\newblock Persformer: 3d lane detection via perspective transformer and the openlane benchmark.
\newblock In \emph{European Conference on Computer Vision}, pages 550--567. Springer, 2022.

\bibitem[Chen et~al.(2023)Chen, Wu, Chitta, Jaeger, Geiger, and Li]{chen2023end}
Li Chen, Penghao Wu, Kashyap Chitta, Bernhard Jaeger, Andreas Geiger, and Hongyang Li.
\newblock End-to-end autonomous driving: Challenges and frontiers.
\newblock \emph{arXiv preprint arXiv:2306.16927}, 2023.

\bibitem[Chen et~al.(2020)Chen, Kornblith, Norouzi, and Hinton]{simclr}
Ting Chen, Simon Kornblith, Mohammad Norouzi, and Geoffrey Hinton.
\newblock A simple framework for contrastive learning of visual representations.
\newblock In \emph{International conference on machine learning}, pages 1597--1607. PMLR, 2020.

\bibitem[Chen et~al.(2017)Chen, Ma, Wan, Li, and Xia]{chen2017multi}
Xiaozhi Chen, Huimin Ma, Ji Wan, Bo Li, and Tian Xia.
\newblock Multi-view 3d object detection network for autonomous driving.
\newblock In \emph{Proceedings of the IEEE conference on Computer Vision and Pattern Recognition}, pages 1907--1915, 2017.

\bibitem[Chitta et~al.(2022)Chitta, Prakash, Jaeger, Yu, Renz, and Geiger]{chitta2022transfuser}
Kashyap Chitta, Aditya Prakash, Bernhard Jaeger, Zehao Yu, Katrin Renz, and Andreas Geiger.
\newblock Transfuser: Imitation with transformer-based sensor fusion for autonomous driving.
\newblock \emph{IEEE Transactions on Pattern Analysis and Machine Intelligence}, 2022.

\bibitem[Gao et~al.(2023)Gao, Geng, Zhang, Ma, Fang, Zhang, Li, and Qiao]{adapter_clip}
Peng Gao, Shijie Geng, Renrui Zhang, Teli Ma, Rongyao Fang, Yongfeng Zhang, Hongsheng Li, and Yu Qiao.
\newblock Clip-adapter: Better vision-language models with feature adapters.
\newblock \emph{International Journal of Computer Vision}, pages 1--15, 2023.

\bibitem[He et~al.(2016)He, Zhang, Ren, and Sun]{resnet}
Kaiming He, Xiangyu Zhang, Shaoqing Ren, and Jian Sun.
\newblock Deep residual learning for image recognition.
\newblock In \emph{Proceedings of the IEEE conference on computer vision and pattern recognition}, pages 770--778, 2016.

\bibitem[Hu et~al.(2021)Hu, Murez, Mohan, Dudas, Hawke, Badrinarayanan, Cipolla, and Kendall]{fiery}
Anthony Hu, Zak Murez, Nikhil Mohan, Sof{\'\i}a Dudas, Jeffrey Hawke, Vijay Badrinarayanan, Roberto Cipolla, and Alex Kendall.
\newblock Fiery: Future instance prediction in bird's-eye view from surround monocular cameras.
\newblock In \emph{Proceedings of the IEEE/CVF International Conference on Computer Vision}, pages 15273--15282, 2021.

\bibitem[Hu et~al.(2022)Hu, Chen, Wu, Li, Yan, and Tao]{STP3}
Shengchao Hu, Li Chen, Penghao Wu, Hongyang Li, Junchi Yan, and Dacheng Tao.
\newblock St-p3: End-to-end vision-based autonomous driving via spatial-temporal feature learning.
\newblock In \emph{European Conference on Computer Vision}, pages 533--549. Springer, 2022.

\bibitem[Hu et~al.(2023)Hu, Yang, Chen, Li, Sima, Zhu, Chai, Du, Lin, Wang, et~al.]{Uniad}
Yihan Hu, Jiazhi Yang, Li Chen, Keyu Li, Chonghao Sima, Xizhou Zhu, Siqi Chai, Senyao Du, Tianwei Lin, Wenhai Wang, et~al.
\newblock Planning-oriented autonomous driving.
\newblock In \emph{Proceedings of the IEEE/CVF Conference on Computer Vision and Pattern Recognition}, pages 17853--17862, 2023.

\bibitem[Jia et~al.(2021)Jia, Yang, Xia, Chen, Parekh, Pham, Le, Sung, Li, and Duerig]{align}
Chao Jia, Yinfei Yang, Ye Xia, Yi-Ting Chen, Zarana Parekh, Hieu Pham, Quoc Le, Yun-Hsuan Sung, Zhen Li, and Tom Duerig.
\newblock Scaling up visual and vision-language representation learning with noisy text supervision.
\newblock In \emph{International conference on machine learning}, pages 4904--4916. PMLR, 2021.

\bibitem[Jiang et~al.(2023)Jiang, Chen, Xu, Liao, Chen, Zhou, Zhang, Liu, Huang, and Wang]{VAD}
Bo Jiang, Shaoyu Chen, Qing Xu, Bencheng Liao, Jiajie Chen, Helong Zhou, Qian Zhang, Wenyu Liu, Chang Huang, and Xinggang Wang.
\newblock Vad: Vectorized scene representation for efficient autonomous driving.
\newblock \emph{arXiv preprint arXiv:2303.12077}, 2023.

\bibitem[Li et~al.(2022)Li, Wang, Li, Xie, Sima, Lu, Qiao, and Dai]{bevformer}
Zhiqi Li, Wenhai Wang, Hongyang Li, Enze Xie, Chonghao Sima, Tong Lu, Yu Qiao, and Jifeng Dai.
\newblock Bevformer: Learning bird’s-eye-view representation from multi-camera images via spatiotemporal transformers.
\newblock In \emph{European conference on computer vision}, pages 1--18. Springer, 2022.

\bibitem[Lin et~al.(2017)Lin, Doll{\'a}r, Girshick, He, Hariharan, and Belongie]{fpn}
Tsung-Yi Lin, Piotr Doll{\'a}r, Ross Girshick, Kaiming He, Bharath Hariharan, and Serge Belongie.
\newblock Feature pyramid networks for object detection.
\newblock In \emph{Proceedings of the IEEE conference on computer vision and pattern recognition}, pages 2117--2125, 2017.

\bibitem[Lin et~al.(2022)Lin, Lin, Pei, Huang, and Su]{sparse4d}
Xuewu Lin, Tianwei Lin, Zixiang Pei, Lichao Huang, and Zhizhong Su.
\newblock Sparse4d: Multi-view 3d object detection with sparse spatial-temporal fusion.
\newblock \emph{arXiv preprint arXiv:2211.10581}, 2022.

\bibitem[Lin et~al.(2023{\natexlab{a}})Lin, Lin, Pei, Huang, and Su]{sparse4dv2}
Xuewu Lin, Tianwei Lin, Zixiang Pei, Lichao Huang, and Zhizhong Su.
\newblock Sparse4d v2: Recurrent temporal fusion with sparse model.
\newblock \emph{arXiv preprint arXiv:2305.14018}, 2023{\natexlab{a}}.

\bibitem[Lin et~al.(2023{\natexlab{b}})Lin, Pei, Lin, Huang, and Su]{sparse4dv3}
Xuewu Lin, Zixiang Pei, Tianwei Lin, Lichao Huang, and Zhizhong Su.
\newblock Sparse4d v3: Advancing end-to-end 3d detection and tracking.
\newblock \emph{arXiv preprint arXiv:2311.11722}, 2023{\natexlab{b}}.

\bibitem[Pan et~al.(2020)Pan, Sun, Leung, Andonian, and Zhou]{pan2020cross}
Bowen Pan, Jiankai Sun, Ho~Yin~Tiga Leung, Alex Andonian, and Bolei Zhou.
\newblock Cross-view semantic segmentation for sensing surroundings.
\newblock \emph{IEEE Robotics and Automation Letters}, 5\penalty0 (3):\penalty0 4867--4873, 2020.

\bibitem[Pan et~al.(2024)Pan, Yaman, Nesti, Mallik, Allievi, Velipasalar, and Ren]{pan2024vlp}
Chenbin Pan, Burhaneddin Yaman, Tommaso Nesti, Abhirup Mallik, Alessandro~G Allievi, Senem Velipasalar, and Liu Ren.
\newblock Vlp: Vision language planning for autonomous driving.
\newblock In \emph{Proceedings of the IEEE/CVF Conference on Computer Vision and Pattern Recognition}, pages 14760--14769, 2024.

\bibitem[Philion and Fidler(2020)]{philion2020lift}
Jonah Philion and Sanja Fidler.
\newblock Lift, splat, shoot: Encoding images from arbitrary camera rigs by implicitly unprojecting to 3d.
\newblock In \emph{Computer Vision--ECCV 2020: 16th European Conference, Glasgow, UK, August 23--28, 2020, Proceedings, Part XIV 16}, pages 194--210. Springer, 2020.

\bibitem[Radford et~al.(2019)Radford, Wu, Child, Luan, Amodei, Sutskever, et~al.]{gpt2}
Alec Radford, Jeffrey Wu, Rewon Child, David Luan, Dario Amodei, Ilya Sutskever, et~al.
\newblock Language models are unsupervised multitask learners.
\newblock \emph{OpenAI blog}, 1\penalty0 (8):\penalty0 9, 2019.

\bibitem[Radford et~al.(2021)Radford, Kim, Hallacy, Ramesh, Goh, Agarwal, Sastry, Askell, Mishkin, Clark, et~al.]{clip}
Alec Radford, Jong~Wook Kim, Chris Hallacy, Aditya Ramesh, Gabriel Goh, Sandhini Agarwal, Girish Sastry, Amanda Askell, Pamela Mishkin, Jack Clark, et~al.
\newblock Learning transferable visual models from natural language supervision.
\newblock In \emph{International conference on machine learning}, pages 8748--8763. PMLR, 2021.

\bibitem[Singh et~al.(2022)Singh, Hu, Goswami, Couairon, Galuba, Rohrbach, and Kiela]{singh2022flava}
Amanpreet Singh, Ronghang Hu, Vedanuj Goswami, Guillaume Couairon, Wojciech Galuba, Marcus Rohrbach, and Douwe Kiela.
\newblock Flava: A foundational language and vision alignment model.
\newblock In \emph{Proceedings of the IEEE/CVF Conference on Computer Vision and Pattern Recognition}, pages 15638--15650, 2022.

\bibitem[Vaswani et~al.(2017)Vaswani, Shazeer, Parmar, Uszkoreit, Jones, Gomez, Kaiser, and Polosukhin]{vaswani2017attention}
Ashish Vaswani, Noam Shazeer, Niki Parmar, Jakob Uszkoreit, Llion Jones, Aidan~N Gomez, {\L}ukasz Kaiser, and Illia Polosukhin.
\newblock Attention is all you need.
\newblock \emph{Advances in neural information processing systems}, 30, 2017.

\bibitem[Wang et~al.(2021{\natexlab{a}})Wang, Xing, and Liu]{clip-4}
Mengmeng Wang, Jiazheng Xing, and Yong Liu.
\newblock Actionclip: A new paradigm for video action recognition.
\newblock \emph{arXiv preprint arXiv:2109.08472}, 2021{\natexlab{a}}.

\bibitem[Wang et~al.(2021{\natexlab{b}})Wang, Zhu, Pang, and Lin]{45_wang2021fcos3d}
Tai Wang, Xinge Zhu, Jiangmiao Pang, and Dahua Lin.
\newblock Fcos3d: Fully convolutional one-stage monocular 3d object detection.
\newblock In \emph{Proceedings of the IEEE/CVF International Conference on Computer Vision}, pages 913--922, 2021{\natexlab{b}}.

\bibitem[Wang et~al.(2021{\natexlab{c}})Wang, Zhu, Pang, and Lin]{fcos3d}
Tai Wang, Xinge Zhu, Jiangmiao Pang, and Dahua Lin.
\newblock Fcos3d: Fully convolutional one-stage monocular 3d object detection.
\newblock In \emph{Proceedings of the IEEE/CVF International Conference on Computer Vision}, pages 913--922, 2021{\natexlab{c}}.

\bibitem[Wang et~al.(2022{\natexlab{a}})Wang, Xinge, Pang, and Lin]{44_wang2022probabilistic}
Tai Wang, ZHU Xinge, Jiangmiao Pang, and Dahua Lin.
\newblock Probabilistic and geometric depth: Detecting objects in perspective.
\newblock In \emph{Conference on Robot Learning}, pages 1475--1485. PMLR, 2022{\natexlab{a}}.

\bibitem[Wang et~al.(2022{\natexlab{b}})Wang, Guizilini, Zhang, Wang, Zhao, and Solomon]{47_wang2022detr3d}
Yue Wang, Vitor~Campagnolo Guizilini, Tianyuan Zhang, Yilun Wang, Hang Zhao, and Justin Solomon.
\newblock Detr3d: 3d object detection from multi-view images via 3d-to-2d queries.
\newblock In \emph{Conference on Robot Learning}, pages 180--191. PMLR, 2022{\natexlab{b}}.

\bibitem[Wortsman et~al.(2022)Wortsman, Ilharco, Kim, Li, Kornblith, Roelofs, Lopes, Hajishirzi, Farhadi, Namkoong, et~al.]{clip-5}
Mitchell Wortsman, Gabriel Ilharco, Jong~Wook Kim, Mike Li, Simon Kornblith, Rebecca Roelofs, Raphael~Gontijo Lopes, Hannaneh Hajishirzi, Ali Farhadi, Hongseok Namkoong, et~al.
\newblock Robust fine-tuning of zero-shot models.
\newblock In \emph{Proceedings of the IEEE/CVF Conference on Computer Vision and Pattern Recognition}, pages 7959--7971, 2022.

\bibitem[Yang et~al.(2023{\natexlab{a}})Yang, Chen, Tian, Tao, Zhu, Zhang, Huang, Li, Qiao, Lu, et~al.]{bevformerv2}
Chenyu Yang, Yuntao Chen, Hao Tian, Chenxin Tao, Xizhou Zhu, Zhaoxiang Zhang, Gao Huang, Hongyang Li, Yu Qiao, Lewei Lu, et~al.
\newblock Bevformer v2: Adapting modern image backbones to bird's-eye-view recognition via perspective supervision.
\newblock In \emph{Proceedings of the IEEE/CVF Conference on Computer Vision and Pattern Recognition}, pages 17830--17839, 2023{\natexlab{a}}.

\bibitem[Yang et~al.(2023{\natexlab{b}})Yang, Chen, Tian, Tao, Zhu, Zhang, Huang, Li, Qiao, Lu, et~al.]{yang2023bevformer}
Chenyu Yang, Yuntao Chen, Hao Tian, Chenxin Tao, Xizhou Zhu, Zhaoxiang Zhang, Gao Huang, Hongyang Li, Yu Qiao, Lewei Lu, et~al.
\newblock Bevformer v2: Adapting modern image backbones to bird's-eye-view recognition via perspective supervision.
\newblock In \emph{Proceedings of the IEEE/CVF Conference on Computer Vision and Pattern Recognition}, pages 17830--17839, 2023{\natexlab{b}}.

\bibitem[Yang et~al.(2021)Yang, Li, Liu, Yu, Ma, He, and Pan]{yang2021projecting}
Weixiang Yang, Qi Li, Wenxi Liu, Yuanlong Yu, Yuexin Ma, Shengfeng He, and Jia Pan.
\newblock Projecting your view attentively: Monocular road scene layout estimation via cross-view transformation.
\newblock In \emph{Proceedings of the IEEE/CVF conference on computer vision and pattern recognition}, pages 15536--15545, 2021.

\bibitem[Yin et~al.(2021)Yin, Zhou, and Krahenbuhl]{52_yin2021center}
Tianwei Yin, Xingyi Zhou, and Philipp Krahenbuhl.
\newblock Center-based 3d object detection and tracking.
\newblock In \emph{Proceedings of the IEEE/CVF conference on computer vision and pattern recognition}, pages 11784--11793, 2021.

\bibitem[Yuan et~al.(2021)Yuan, Chen, Chen, Codella, Dai, Gao, Hu, Huang, Li, Li, et~al.]{yuan2021florence}
Lu Yuan, Dongdong Chen, Yi-Ling Chen, Noel Codella, Xiyang Dai, Jianfeng Gao, Houdong Hu, Xuedong Huang, Boxin Li, Chunyuan Li, et~al.
\newblock Florence: A new foundation model for computer vision.
\newblock \emph{arXiv preprint arXiv:2111.11432}, 2021.

\bibitem[Zhang et~al.(2021)Zhang, Fang, Zhang, Gao, Li, Dai, Qiao, and Li]{clip-2}
Renrui Zhang, Rongyao Fang, Wei Zhang, Peng Gao, Kunchang Li, Jifeng Dai, Yu Qiao, and Hongsheng Li.
\newblock Tip-adapter: Training-free clip-adapter for better vision-language modeling.
\newblock \emph{arXiv preprint arXiv:2111.03930}, 2021.

\bibitem[Zhang et~al.(2022{\natexlab{a}})Zhang, Jiang, Miura, Manning, and Langlotz]{zhang2022contrastive}
Yuhao Zhang, Hang Jiang, Yasuhide Miura, Christopher~D Manning, and Curtis~P Langlotz.
\newblock Contrastive learning of medical visual representations from paired images and text.
\newblock In \emph{Machine Learning for Healthcare Conference}, pages 2--25. PMLR, 2022{\natexlab{a}}.

\bibitem[Zhang et~al.(2022{\natexlab{b}})Zhang, Zhu, Zheng, Huang, Huang, Zhou, and Lu]{beverse}
Yunpeng Zhang, Zheng Zhu, Wenzhao Zheng, Junjie Huang, Guan Huang, Jie Zhou, and Jiwen Lu.
\newblock Beverse: Unified perception and prediction in birds-eye-view for vision-centric autonomous driving.
\newblock \emph{arXiv preprint arXiv:2205.09743}, 2022{\natexlab{b}}.

\bibitem[Zhou et~al.(2022{\natexlab{a}})Zhou, Yang, Loy, and Liu]{clip-1}
Kaiyang Zhou, Jingkang Yang, Chen~Change Loy, and Ziwei Liu.
\newblock Learning to prompt for vision-language models.
\newblock \emph{International Journal of Computer Vision}, 130\penalty0 (9):\penalty0 2337--2348, 2022{\natexlab{a}}.

\bibitem[Zhou et~al.(2022{\natexlab{b}})Zhou, Yang, Loy, and Liu]{clip-3}
Kaiyang Zhou, Jingkang Yang, Chen~Change Loy, and Ziwei Liu.
\newblock Conditional prompt learning for vision-language models.
\newblock In \emph{Proceedings of the IEEE/CVF Conference on Computer Vision and Pattern Recognition}, pages 16816--16825, 2022{\natexlab{b}}.

\bibitem[Zhou et~al.(2022{\natexlab{c}})Zhou, Yang, Loy, and Liu]{coop}
Kaiyang Zhou, Jingkang Yang, Chen~Change Loy, and Ziwei Liu.
\newblock Learning to prompt for vision-language models.
\newblock \emph{International Journal of Computer Vision}, 130\penalty0 (9):\penalty0 2337--2348, 2022{\natexlab{c}}.

\bibitem[Zhu et~al.(2020)Zhu, Ma, Wang, Xu, Shi, and Lin]{55_zhu2020ssn}
Xinge Zhu, Yuexin Ma, Tai Wang, Yan Xu, Jianping Shi, and Dahua Lin.
\newblock Ssn: Shape signature networks for multi-class object detection from point clouds.
\newblock In \emph{Computer Vision--ECCV 2020: 16th European Conference, Glasgow, UK, August 23--28, 2020, Proceedings, Part XXV 16}, pages 581--597. Springer, 2020.

\end{thebibliography}
}


\end{document}